\documentclass{article}

\usepackage{microtype}
\usepackage{graphicx}
\usepackage{subfigure}
\usepackage{booktabs} % for professional tables
\usepackage{hyperref}
\usepackage{bbm}
\usepackage{amsmath}
\usepackage{amssymb}
\usepackage{bm}

\usepackage[accepted]{icml2022}
\usepackage{color}         % colors
\usepackage{amsmath} %
\newcommand{\vect}[1]{\boldsymbol{#1}}

\newcommand{\rev}[2][]{\textcolor{red}{\sout{#1}}\textcolor{blue}{#2}} % revision
\newcommand{\note}[2]{\textbf{#1} (\textcolor{blue}{#2})} % note for comments
\newcommand{\para}[1]{\noindent\textbf{#1}\quad} % The topic of a paragraph

\newcommand{\commandusage}[1]{
\para{Usage of Commands} \note{This paragraph shows how to use these commands.}{Uncomment commandusage before submitting}
\rev[This is a wrong sentence.]{This is a right sentence.}
}

\icmltitlerunning{Object-based Masked Autoencoders for Fast Pre-training}

\begin{document}

\twocolumn[
\icmltitle{Object-wise Masked Autoencoders for Fast Pre-training}




\begin{icmlauthorlist}
\icmlauthor{Jiantao Wu}{uj}
\icmlauthor{Shentong Mo}{cmu}
\end{icmlauthorlist}

\icmlaffiliation{uj}{University of Jinan}
\icmlaffiliation{cmu}{Carnegie Mellon University}
% \icmlaffiliation{ed}{School of Computation, University of Edenborrow, Edenborrow, United Kingdom}

\icmlcorrespondingauthor{Jiantao Wu}{530896890@qq.com}
% \icmlcorrespondingauthor{Eee Pppp}{ep@eden.co.uk}

\icmlkeywords{Machine Learning, ICML}

\vskip 0.3in
]



\printAffiliationsAndNotice{} % otherwise use the standard text.

\begin{abstract}
Self-supervised pre-training for images without labels has recently achieved promising performance in image classification.
The success of transformer-based methods, ViT and MAE, draws the community's attention to the design of backbone architecture and self-supervised task. 
In this work, we show that current masked image encoding models learn the underlying relationship between all objects in the whole scene, instead of a single object representation. 
Therefore, those methods bring a lot of compute time for self-supervised pre-training.
To solve this issue, we introduce a novel object selection and division strategy to drop non-object patches for learning object-wise representations by selective reconstruction with interested region masks.
We refer to this method \textbf{ObjMAE}. 
Extensive experiments on four commonly-used datasets demonstrate the effectiveness of our model in reducing the compute cost by 72\% while achieving competitive performance. 
Furthermore, we investigate the inter-object and intra-object relationship and find that the latter is crucial for self-supervised pre-training.

\end{abstract}

\begin{figure}[!htb]
    \centering
    \includegraphics[width=\linewidth]{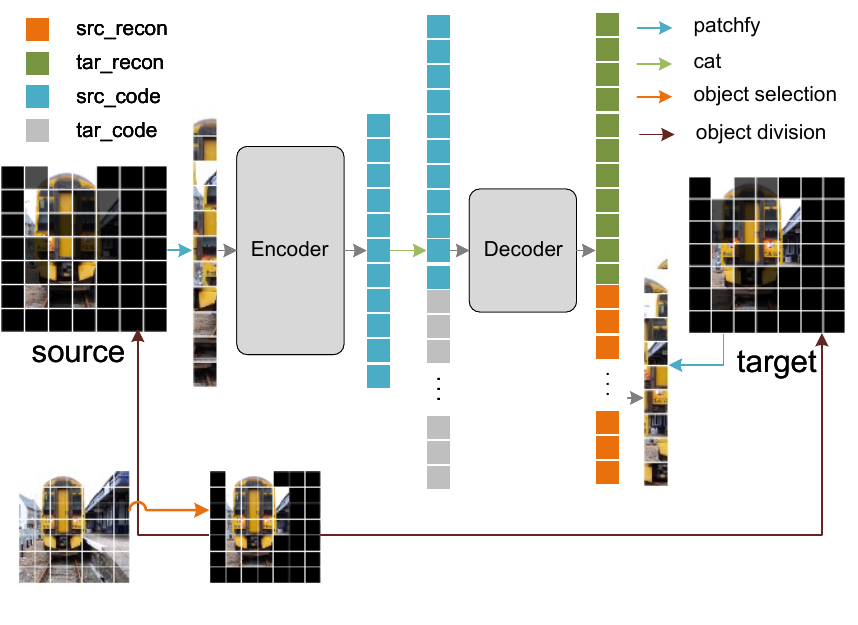}
    \vspace{-3em}
    \caption{Overview of our ObjMAE based on MAE~\cite{he2021masked} for fast pre-training. 
    ObjMAE select one object and divide it into a source region, a target region, and a discarded region, differing with MAE in the selection from all objects. 
    The encoder takes source patches with selective object masks as input, and the decoder predicts the target patches of objects.
    ObjMAE has a light self-supervised prediction task for only the object region and ignores the inter-object semantics. 
    Note that \textbf{BLACK} patches are discarded.
    } \label{fig:self-supervised_task}
\end{figure}

\section{Introduction}
Understanding the objects in a complex scene is a fundamental ability of human intelligence, which is abstracted by computer experts as classification.
However, the traditional classification setup by supervision requires training with numerous labels for the target objects, unlike humans, who only need few supervision information.
To overcome this gap, self-supervised pre-training~\citep{erhan2010does,dosovitskiy2014discriminative, wu2018unsupervised,chen2020simple,chen2020big,grill2020bootstrap,he20moco,chen2020mocov2,chen2021simsiam,jure2021barlow} is a promising approach to enable models to learn without expensive labels.
Until recently, the self-supervised pre-trained models~\citep{caron2021emerging,xie2021self-supervised,he2021masked,bao2021beit} surpass the supervised models on computer vision tasks with the help of transformer architectures~\citep{ViT,touvron2020deit,liu2021Swin,yuan2021tokens,ElNouby2021XCiTCI}.

Typically, Masked Autoencoders (MAE)~\citep{he2021masked} suggests that it is a good self-supervised task to randomly mask 75\% of patches in an image.
The model shows a strong ability to rebuild the original image or in-paint the missing parts.
In other words, there is much redundant information on an image, and 25\% is enough to represent it.
MAE promotes the model to focus on the valuable parts by throwing redundant information, thereby extracting a better representation with higher-level abstract.

\begin{figure}[!htb]
    \centering
    \includegraphics[width=\linewidth]{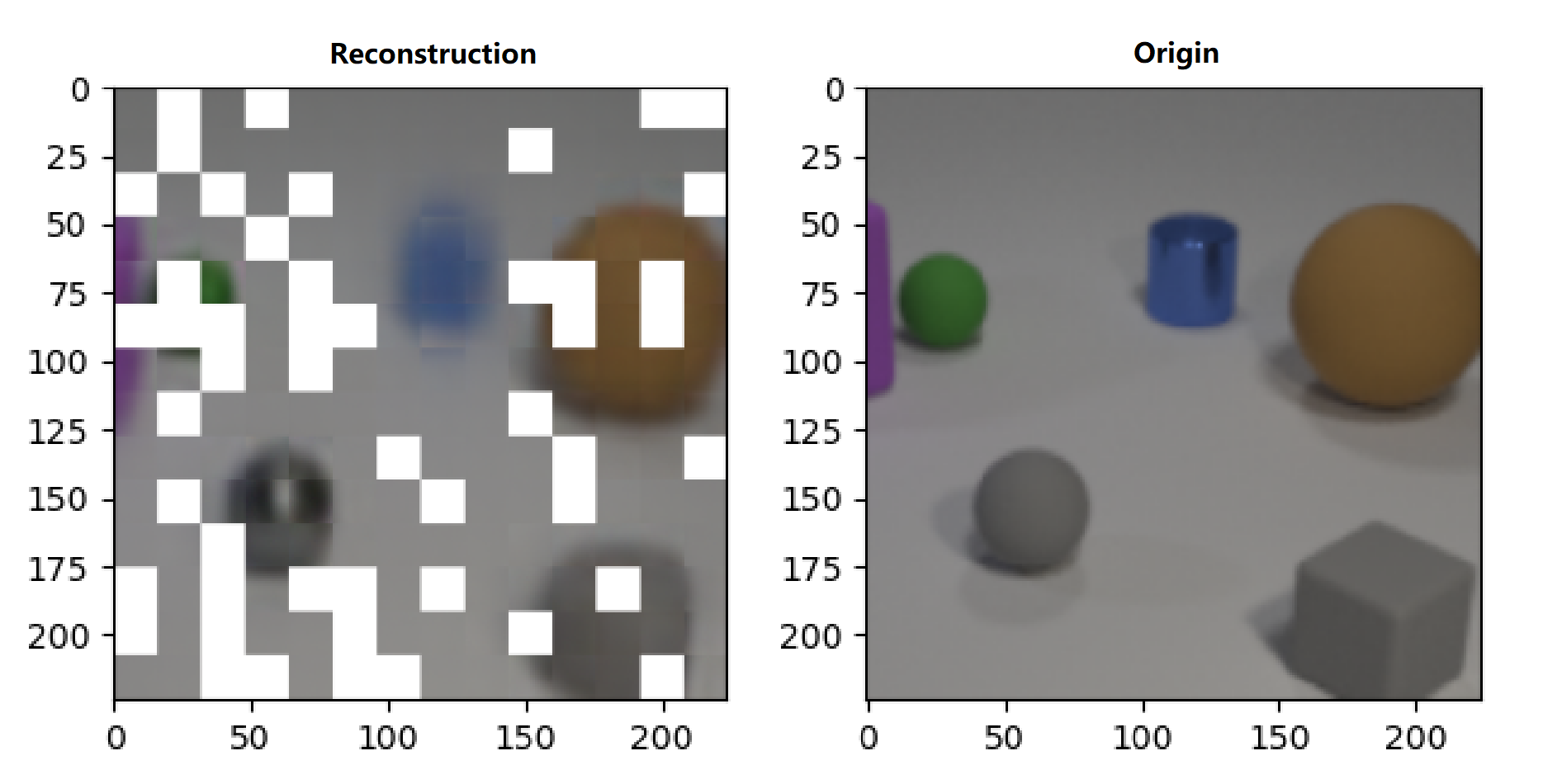}
    \vspace{-0.4cm}
    \caption{The reconstruction result (left) of MAE~\citep{he2021masked} for the masked image (right).
    We pre-train a standard MAE with a masking ratio of 0.75 on CLEVR-M for 300 epochs. 
    The white patches on the left are the visible patches, and the others are the reconstructed patches for the target region.
    Note that, all patches of the \textbf{BLUE} cylinder are invisible for MAE, yet MAE rebuilds the unseen object.
    MAE memorizes this image in the pre-training iterations and retrieves it according to the inter-object information, like positions.
    Unlike our propose ObjMAE, MAE performs poorly to learn object-wise representations for self-supervised training.
    }
    \label{fig:mae_recon}
\end{figure}

We notice that MAE can rebuild the unseen object, indicating overfitting or memorizing the image as shown in Figure~\ref{fig:mae_recon}.
Therefore, the model learns an image-wise representation may contain several objects. 
However, humans rarely regard one picture as a whole thing. 
Instead, it is more natural to recognize each small part in the picture and process information extracted from the objects.
A question comes to mind: \textbf{will the pre-training model benefit from an object-wise representation?}

Our goal is to propose an object-wise representation learning method where the extracted features of a patch only contain information belonging to the same object.
We hypothesize that an object-wise representation promotes the performance of the pre-training model by decoupling the biases among objects.
For instance, birds and the sky are likely to appear in couple, and the model may mistakenly recognize a flying or a bird based on the sky consequently.

To obtain an object-wise representation, we propose a novel object selection and division strategy based on Masked Autoencoder, which we call \textbf{ObjMAE}.
Our ObjMAE prevents the model from learning the inter-object relationship by only predicting the patches belonging to the sample object of the visible patches in the training process.
Furthermore, our method accelerates the pre-training process by throwing many patches and reduces the compute cost by 72\% on MS-COCO~\citep{COCO} and ImageNet-100~\citep{imagenet}.
We also demonstrate the competitive transfer performance of our model on CIFAR100 and ImageNet-100.

Experiments on CLEVR examine the potency of object-wise representations on reasoning and show that the intra-object prediction is more efficient for self-supervised pre-training.

Overall, the contributions of our work are summarized as follows:
\begin{itemize}
    \item We propose an object-wise masked autoencoder named ObjMAE with a novel selective masking strategy.
    \item We demonstrate the effectiveness of our method in reducing the compute cost of pre-training by 72\% and achieving competitive transfer performance.
    \item We also explore the inter-object and intra-object relationship in an image and find that the latter is crucial for self-supervised pre-training.
\end{itemize}

\section{Related Work}

\subsection{Masked Image Modeling}

Masked image encoding has shown its promising performance in unsupervised representation learning for the whole image. Typically, BEiT~\citep{bao2021beit} leverages the block-wise masking strategy on image patches and recovers the discrete tokens of masked patches during pre-training. 
Following BEiT, SimMLM~\citep{xie2021SimMIM} proposes to use a large randomly masked patch size and strong pre-text task for regressing raw pixels of RGB values with a lighter prediction head. 
A better perceptual codebook in PeCo~\citep{dong2021peco} enforces the perceptual similarity during the dVAE~\citep{ramesh2021zero-shot} training to learn better semantic meanings for achieving superior performance in downstream tasks. 
MAE~\citep{he2021masked} applies an encoder and a lightweight decoder to reconstruct the missing pixels by masking random patches of the input image. 
MaskFeat~\citep{wei2021masked} uses Histograms of Oriented Gradients (HOG) as the pre-training target of the masked regions in unlabeled videos.
More recently, SplitMask~\citep{alaaeldin2021are} conducts a self-supervised pre-training paradigm with the target task data to obtain better performance on the COCO dataset. 
iBot performs masked prediction with a teacher network as an online tokenizer.
However, in this work, we focus on learning representation from the object-level patch and introduce an object-wise masked autoencoder with a novel mask strategy. Our method only predicts the patches belonging to the sample object of the visible patches in the training process, while accelerating the pre-training process by throwing many redundant patches.

\subsection{Object-wise Representations Learning}

Reasoning about discrete objects in a scene is foundational for humans to understand the world.
A wide range of works has successfully developed unsupervised segmentation by inferring the objects in images~\citep{Martin21GENESIS-V2,huang2015efficient}.
These methods learn intra-object semantics to identify the difference between objects.
Previous works only explore the application of object-wise representation in segmentation.
In this work, we explore it in self-supervised pre-training for image classification.

\begin{figure*}[!htb]
    \centering
    \includegraphics[width=.9\linewidth]{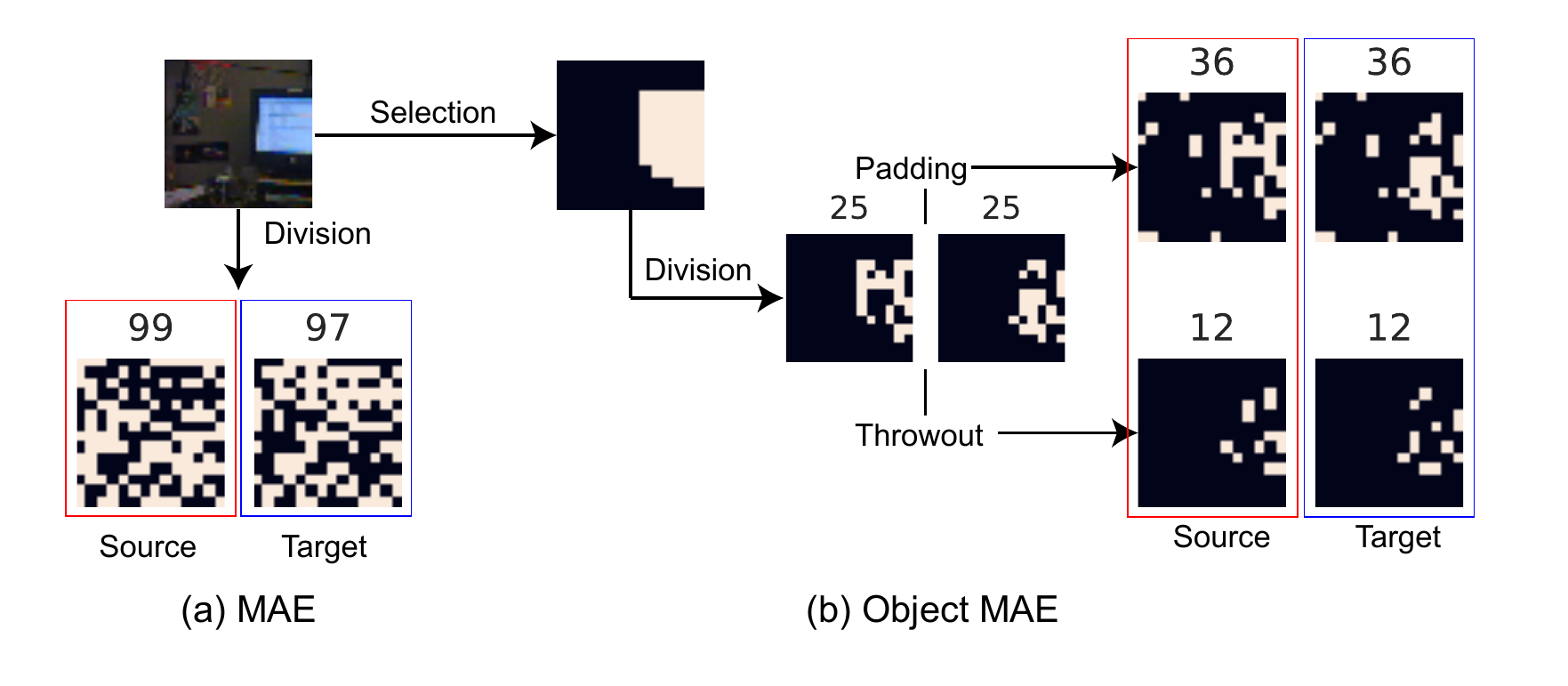}
    \vspace{-0.7cm}
    \caption{Comparison of source patches and target patches between MAE and ObjMAE. The source patches are highlighted in red box and the target patches are highlighted in blue box. (a) MAE select patches from the whole image. (b) The object region is split into  a sequence of patches and then divided into source patches and target patches. Our method has three steps: selection, division, alignment. 
    }
    \label{fig:method}
\end{figure*}

\section{Method}

\subsection{ObjMAE}
The conventional vision transformer models reshape the 2D image into a sequence of patches $\vect{x}_p\in \mathbb{R}^{(H W) \times (P^2 C)}$ and utilize all of them,  where $(H P, W P)$ is the resolution of the original image, $C$ is the number of channels, $(P, P)$ is the resolution of each image patch. 
Instead, we only use a subset of $\vect{x}_p^s$ from $\vect{x}_p$ and separate the selective 2D patches $\vect{x}_p^s$ into three parts: the source patches $\mathcal{X}^{\mathrm{src}}$, the target patches $\mathcal{X}^{\mathrm{tar}}$, and the discarded patches $\mathcal{X}^{\mathrm{out}}$.
The source patches are the input of our model, and our model tries to rebuild the contents of the target token in terms of the source token.
The discarded region will not be used in all procedures, which reduces the computational cost.

An overview of patch division is depicted in Figure~\ref{fig:method}. 
First, we randomly choose an object in a scene and select all patches belonging to the object.
Then, we randomly divide the object region into the source token and the target token equally.
Finally, we align the sequences of source tokens and target tokens with a fixed size according to a hyperparameter $\alpha$.

The main difference between MAE and ObjMAE is the selection process where MAE picks patches from the whole image instead of the object region.
Our model equals to MAE when the source token and the target token are randomly picked from the whole image, and there is no discarded region.

\subsection{Object Selection}
The segmentation information of objects is not available for most classification tasks.
One solution is to pre-train our model on a dataset with segmentation.
Another solution is to use an extra model to detect the interested  region for objects.

\paragraph{Selection with Segmentation.} 
This method selects an object according to the labels of image segmentation $y_p\in \mathcal{R}^{(H P)\times(W P)}$.
Note that, we do not need to know the category of the segmentation.
In other words, the cost of collecting such segmentation is cheaper than the normal one.
To ensure each object is selected impartially, we uniformly sample an object on an image.
To avoid missing information of object, we regard a patch as a part of an object if 20\% pixels belong to it.

\paragraph{Selection with CAM.}
Class Activation Mapping~\citep{cam} is a popular technique to identify the interested region of a deep neural network for a given class.
We use CAM to capture a rough region for the object on an image here. 
We apply a pre-trained ResNet-50~\citep{resnet} on ImageNet-1K~\citep{imagenet} to obtain an object region roughly, see details in Appendix A.

\subsection{Object Division \& Alignment}
Each object region will be divided into a source token---the visible part of the encoder and a target token---the rebuilt part of the decoder.
In this work, we divide an object region equally.
It is equal to the case when the masking ratio is 0.5 in MAE.
MAE has shown the effect of masking ratio, and it is not the point that this work aims to.

\begin{figure*}[!htb]
    \centering
    \subfigure[CLEVR]{\includegraphics[width=.49\linewidth]{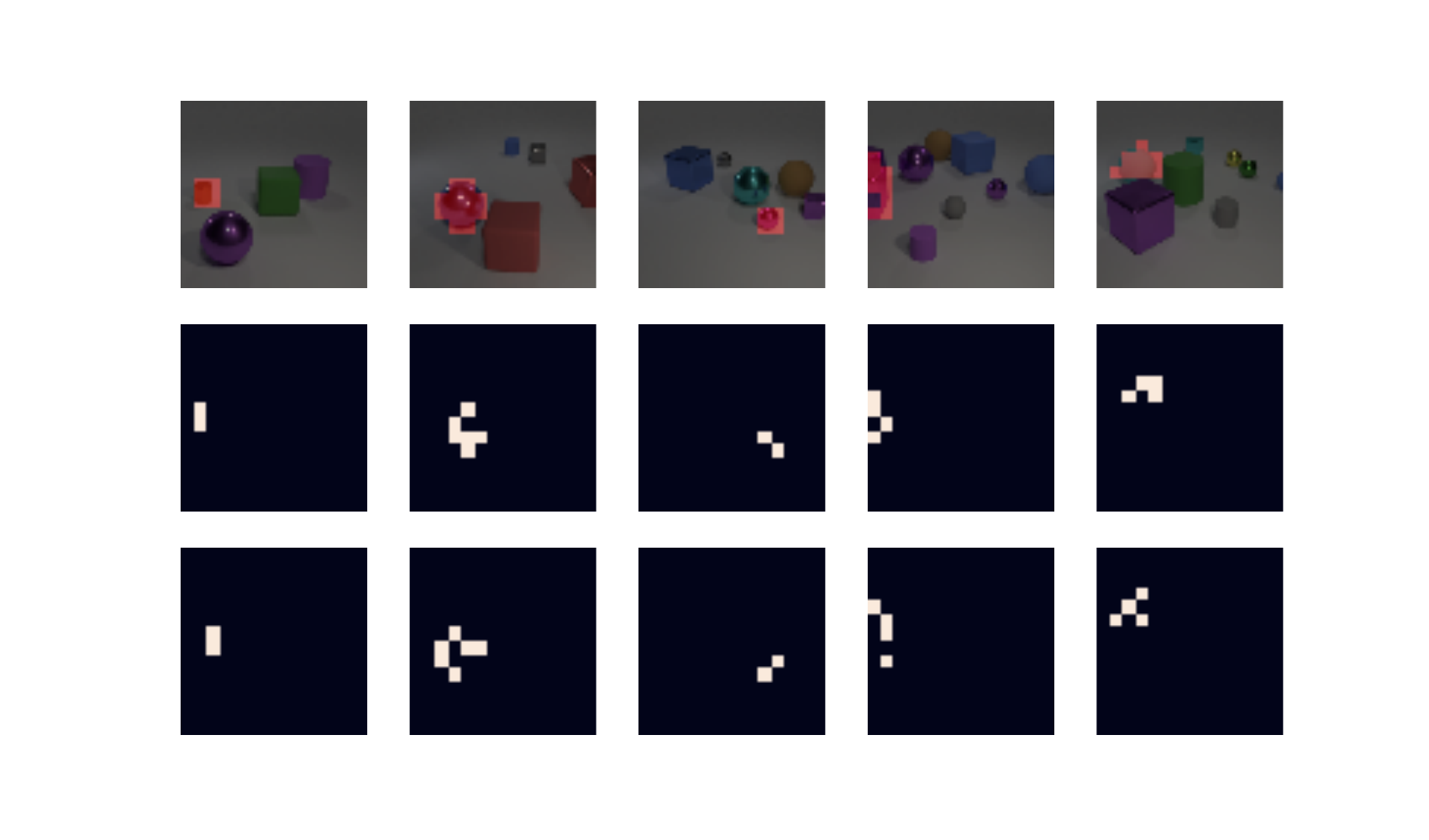}}
    \subfigure[COCO]{\includegraphics[width=.49\linewidth]{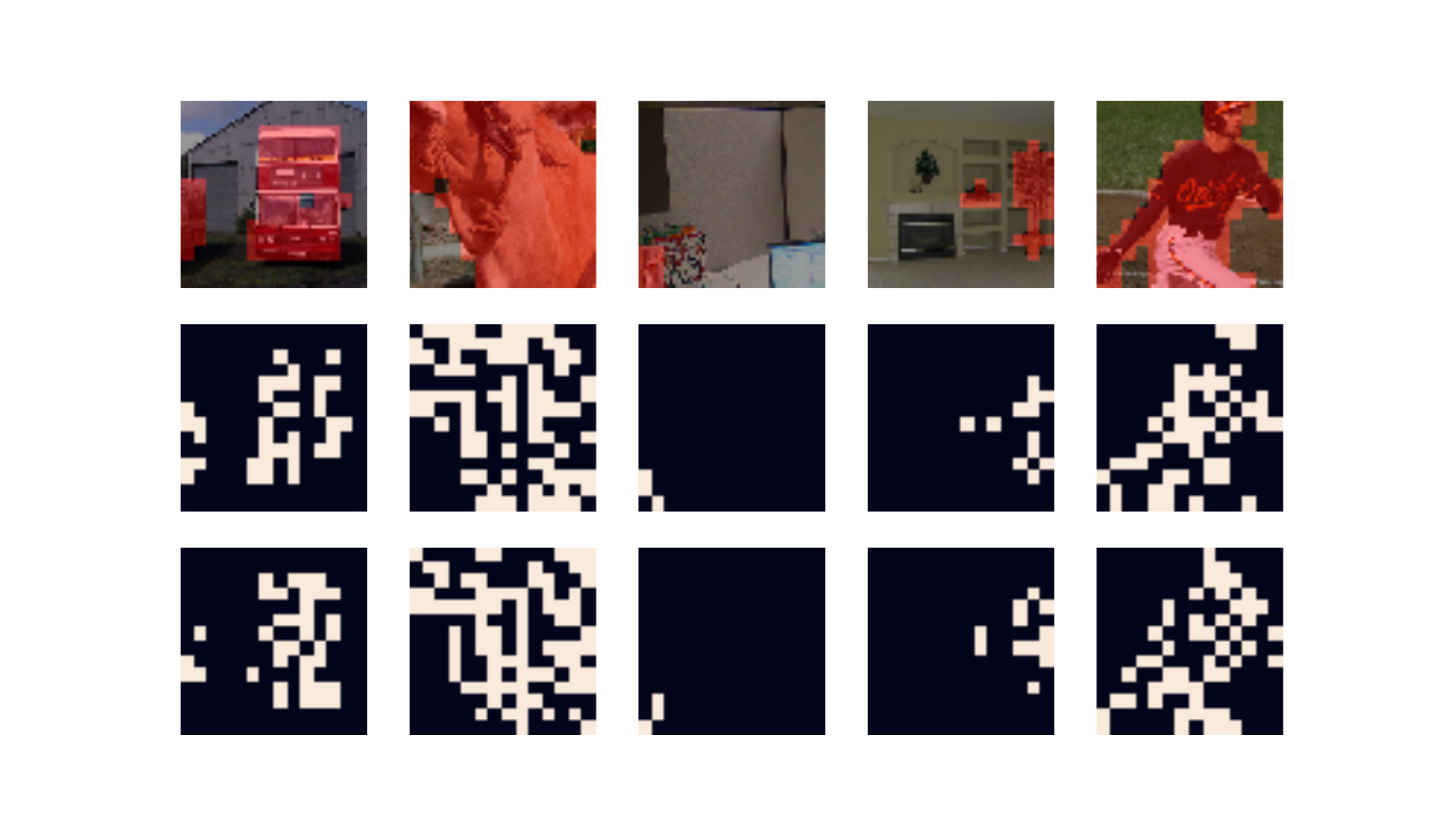}} %
    \caption{Dataset Visualization. The first row shows the image with the object region denoted by red. The second row and third row are the target region and the source region respectively.}
    \label{fig:dataset}
\end{figure*}

The sizes of objects on the data are various, causing difficulty in training in a mini-batch with different sizes.
It can be solved with deliberate programming ideally, but it is too costly in the current deep learning framework, like PyTorch.
Therefore, we introduce a hyperparameter $\alpha$ to control the patches of input on a batch.
The lengths of both the source sequences and target sequences are $\alpha H W$.

\paragraph{Throwout.} 
Both source tokens and target tokens are aligned to the region with the lowest size in one mini-batch.
Specifically, the regions with larger sizes will be cut by randomly throwing more patches.
This strategy causes fewer patches will be calculated, which may lose the information of the object.

\paragraph{Zero Padding.} 
Zero Padding is simply a process of adding patches to the source sequence or the target sequence so as to align the length of the input.
The added patches are filled with zero so that they contain no information.

\paragraph{Replicate Padding.} 
We will repeat the patches on the sequence until the length meets the required number.
The added patches are repeated so that no extra information is introduced.

\paragraph{Random Padding.} 
We randomly add patches from all patches.
This method has the risk of incorporating information from other objects or the background.

\subsection{Self-supervised Task}

MAE can be seen as a special case of ObjMAE, and the implementation is similar.
The self-supervised task is demonstrated in Figure~\ref{fig:self-supervised_task}.
ObjMAE reconstructs the target patches according to the source patches.
First, we generate source tokens from source patches by linear projection with added positional embeddings. 
After encoding, we append a list of mask tokens with positional embeddings of target patches to the list of encoded patches.
Then, the decoder reconstructs the original target region in pixels.
Unlike MAE~\citep{he2021masked} that \textit{shuffles} or \textit{unshuffles} the token list, we \textit{gather} the vector of target patches in the correct order.
Our loss function computes the mean squared error (MSE) between the reconstructed and original images of target parches in the pixel space.

\begin{figure*}[!htb]
    \centering
    \subfigure[CLEVR ]{\includegraphics[width=.5\linewidth]{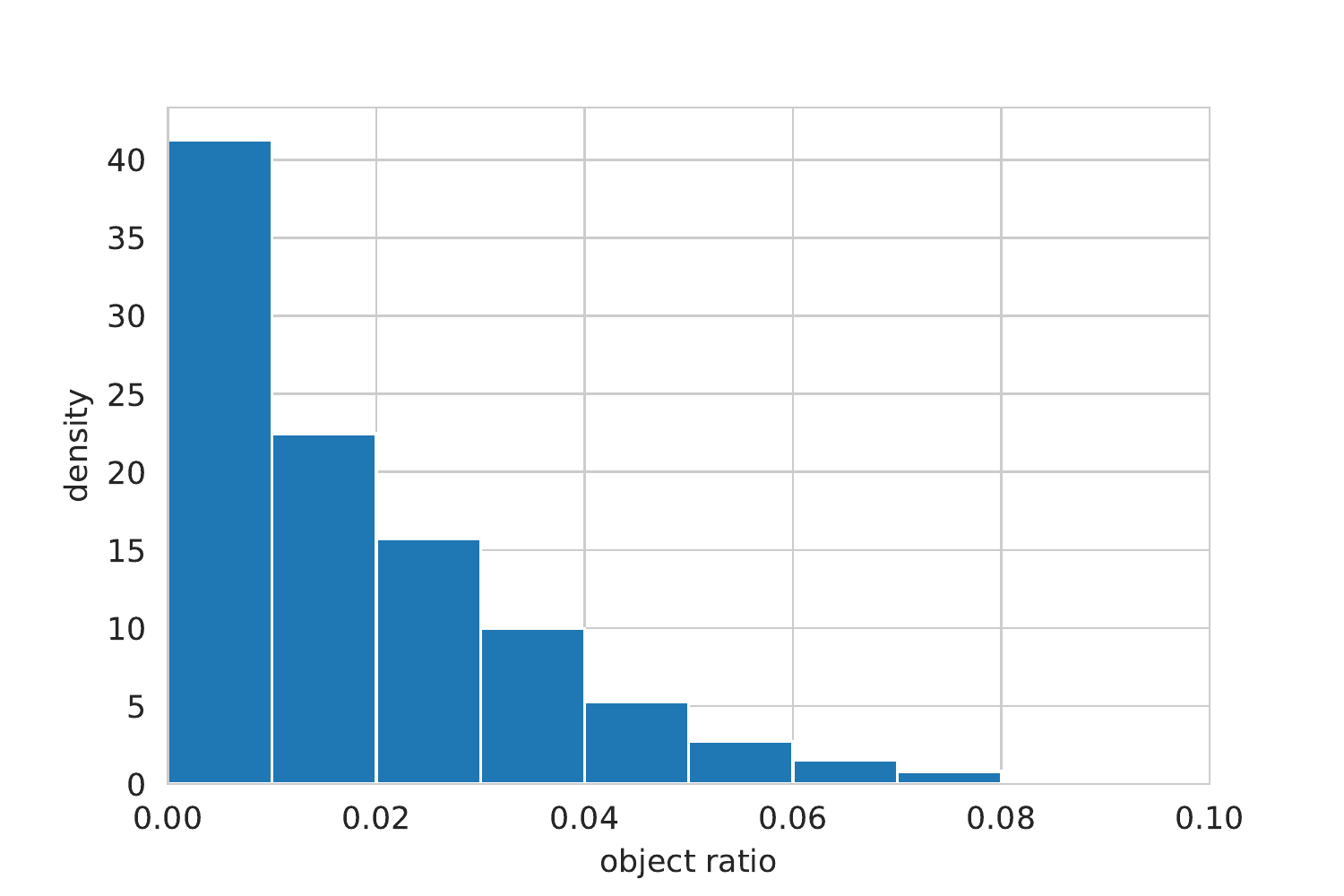}}%
    \subfigure[COCO]{\includegraphics[width=.5\linewidth]{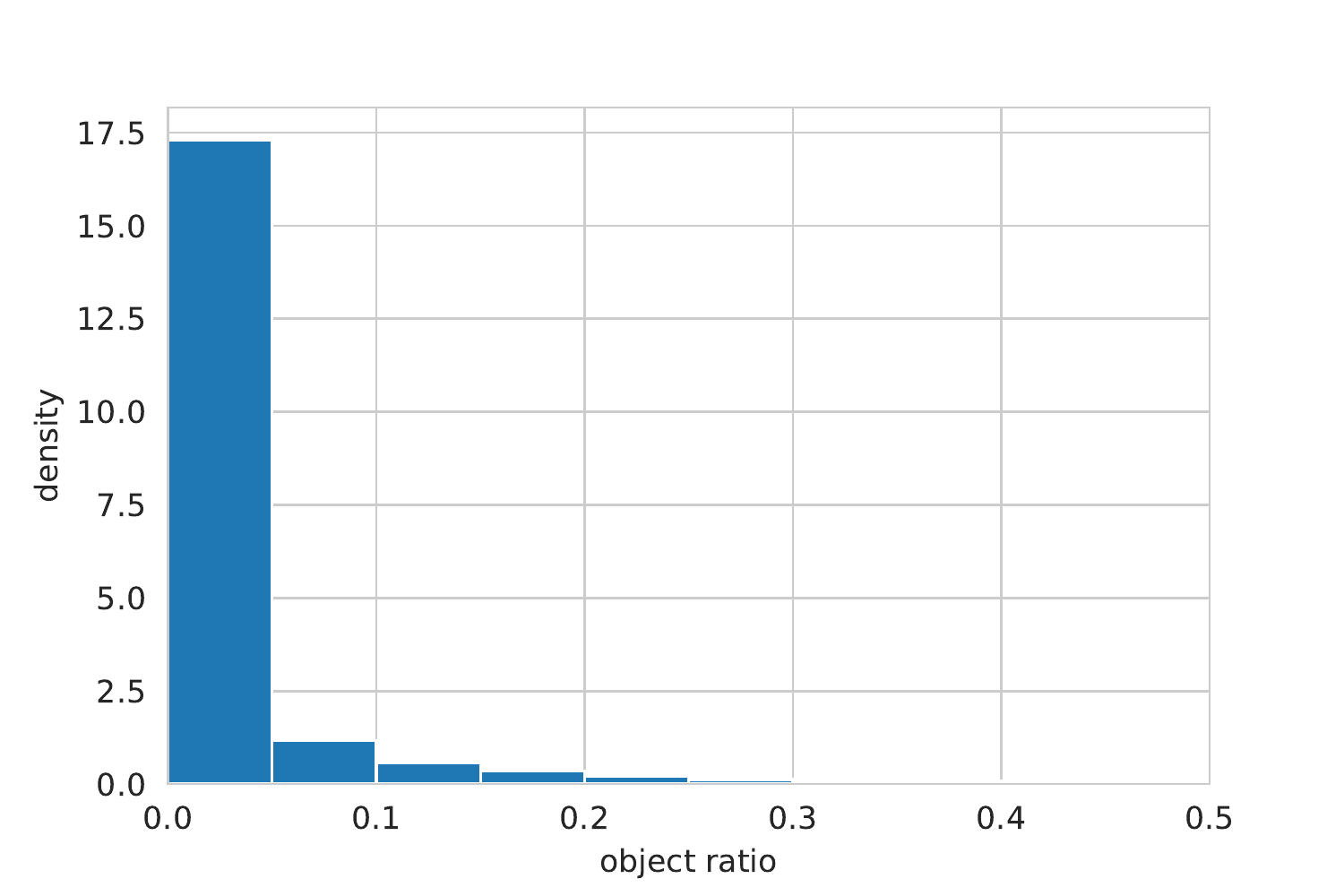}}
    \vspace{-1em}
    \caption{The distributions of size ratio on CLEVR and COCO. 70\% objects are smaller than 0.1. 99\% objects are smaller than 0.1. }
    \label{fig:obj_size}
\end{figure*}

\section{Experiments}

\subsection{Datasets}

\paragraph{MS-COCO.} MS-COCO~\citep{COCO} is a large-scale dataset for object detection, segmentation, and captioning. 
COCO provides precious instance label information for every pixel.
The instance label helps us to select the patches belonging to the same object.
We get a cleaner dataset with 72, 452 samples by dropping those with a small size or containing no object from the training split of COCO2014 with 82, 783 samples.
The average object ratio is 0.34, so we set $\alpha$ to 0.34.
We study the effects of inter-object semantics on this dataset with accurate segmentation.

\paragraph{ImageNet-100.} ImageNet-100 is a subset of ImageNet-1k~\citep{imagenet}, containing 100 classes. 
There are about 1,300 samples for each class.
ImageNet-1k is a popular benchmark on computer vision, and many works are based on it.
Here, we use the ResNet-34~\citep{resnet} pre-trained on ImageNet-1K to get saliency maps. 
ImageNet-100 is the benchmark to evaluate the performance of pre-trained models. 

\paragraph{CIFAR-100.} CIFAR-100~\citep{krizhevsky2009learning} is a popular benchmark for classification with 100 classes and 60k images. 
We resize the images on CIFAR-100 from 32x32 to 224x224.
This dataset is only used to fine-tune the pre-trained models. 

\paragraph{CLEVR.} CLEVR\citep{johnson2017clevr} is a popular artificial dataset for visual reasoning. 
There is no underlying relationship between objects---all objects are generated independently.
To get the instance information, we modified the code~\footnote{\href{CLEVR Dataset Generation}{https://github.com/facebookresearch/clevr-dataset-gen}} and generated 50K samples with instance labels (40K for training, 10K for test), namely CLEVR-M.
The average object ratio of objects on CLEVR-M is 0.05.
We visualize the learned representations of MAE and ObjMAE on this dataset and identify the shortage of inter-object semantics.

\paragraph{Dataset Visualization.}
We show the samples of CLEVR and COCO and their object segmentation in Figure~\ref{fig:dataset}.

\paragraph{Object Ratio.}
The calculation of object ratio is:
\begin{equation}
    S(\vect{y}=i)/S(\vect{y}),
\end{equation}
where $S$ counts the number of pixels, $\vect{y}$ is segmentation, $i$ is the index of selected object.

We show the distribution of object ratio in Figure~\ref{fig:obj_size}.
One can see that most objects are extremely small (lower than 0.05\%), which means that the object-wise prediction task saves a lot computational time.
Note that, we discard the objects having object ratio smaller than 0.05 for pre-training. 
As a result, the average object ratio is 0.05 for CLEVR-M, 0.41 for ImageNet-100, and 0.34 for COCO.

\begin{table}[]
    \centering
    \begin{tabular}{ccccc}
    \toprule
                & Layers & $D$  & Heads & Params\\
       \midrule
      Encoder &  12 & 384 & 6 & 21M\\
      \midrule
      Decoder & 4 & 192 & 3 & 19M\\
      \bottomrule
    \end{tabular}
    \caption{Details of our model. $D$ denotes Hidden Size.}\label{tab:architecture}
\end{table}

% \usepackage{booktabs}
\begin{table}[!htb]
\centering
\scalebox{0.9}{
\begin{tabular}{@{}l|c@{}}
\toprule
Hyperparameter                 &    Value                 \\ \midrule
Optimizer                      & AdamW, Lookahead        \\
AdamW $\epsilon$               & le-8                     \\
AdamW $\beta$                  & (0.9,0.999)              \\
Weight decay                   & 0.05 \\
Layer-wise learning rate decay & 0.65                     \\ \midrule
Learning rate schedule         & Cosine                   \\
Minimal learning rate          & 1e-5                     \\ 
% Learning rate                   & 1.5e-4                  \\ \midrule
% Dropout                        & 0                        \\
% Batch size                     & 256                      \\
Input resolution               & 224×224                  \\
Position embedding interpolate & Sinusoid                 \\ \bottomrule
\end{tabular}}
\caption{Default hyperparameters. Optimizer is AdamW~\citep{adamw} wrapped with Look-ahead~\citep{lookahead}.}\label{tab:hyperparameter}
\end{table}

\begin{table}[!htb]
\centering
\scalebox{0.9}{
\begin{tabular}{lllll}
\toprule
Dataset     & Epoch & Warm-up & BS & LR \\ \midrule
\multicolumn{5}{c}{Pre-training}          \\ \midrule
CLEVR       & 300         & 40      & 256 &1.5e-4\\
COCO        & 800         & 40      & 128 &1.5e-4\\
ImageNet-100 & 800         & 40      & 128 &1.5e-4\\ \midrule
\multicolumn{5}{c}{Finetune}              \\ \midrule
ImageNet-100 & 100         & 5       & 256 &1e-3\\
CIFAR100    & 90          & 5       & 256 &1e-3\\ \bottomrule
\end{tabular}}
\caption{Training settings for datasets. BS denotes batch size. LR denotes learning rate.}\label{tab:training_settings}
\end{table}

\subsection{Experimental Setup}
\paragraph{Pre-processing.} For the pre-training process, all images are randomly cropped into 224x224 and then normalized with the mean and standard deviation from ImageNet-1K.
For the fine-tuning process, we follow the same settings in~\citep{KolesnikovBZPYG20}. 

\paragraph{Architecture.}
We reduce the hidden size of ViT-B~\citep{ViT} and obtain ViT-Small as shown in Table~\ref{tab:architecture}, namely ViT-S.
By default, the patch size is 16 for all experiments.
Compared with MAE, the input of the decoder is a subset of patches---the target patches and source patches.

\paragraph{Training Setting.}
The general settings for both pre-training and fine-tuning are listed in Table~\ref{tab:hyperparameter}.
Table~\ref{tab:training_settings} shows the particular settings for tasks.

\subsection{Pre-training Acceleration}
% \usepackage{booktabs}
% \usepackage{booktabs}
\begin{table}[]
\centering
\begin{tabular}{@{}c|llllll@{}}
\toprule
$\alpha$     & Speed Up &Recon Loss & Top-1     & Top-5\\ \midrule
1/2          & 1.9x        & 0.5603    & \textbf{89.52}     & \textbf{98.16} \\
\textbf{1/4} & \textbf{3.6x}       & \textbf{0.4874}    & 88.69   & 98.06    \\
1/8          & 6.4x     &0.6665    &    84.74   & 94.60   \\ 
1/16          & 9.2x       &0.6995    &    84.58   & 94.14   \\ \midrule
\end{tabular}
\caption{Performance vs. acceleration. Speed Up is compared with standard ViT-S on a mini-batch with 256 samples in a single machine with one TITAN RTX. Recon Loss denotes the loss of predicting target patches for pre-training. Top-1 and Top-5 are evaluated on ImageNet-100 for fine-tuning. $\alpha$ equals to MAE. ObjMAE achieves comparable performances within 27\% runtime when $\alpha=0.25$}\label{tab:acceleration}
\end{table}
In this part, we limit the selective region to reduce the computational consumption by controlling $\alpha$.
We use CAM~\citep{cam} to get saliency maps of images on ImageNet-100.
Then, we select the patches with the most responsible value.

We train the ObjMAE on ImageNet-100 and evaluate on ImageNet-100.
We compare the performance over $\alpha$ in Table.~\ref{tab:acceleration}.
Though MAE has a relevant high Recon Loss, it has the highest performance.
The model can hardly predict the target regions correctly as $\alpha$ goes down (Recon loss increases), resulting in poorer performances.
ObjMAE achieves competitive performance by throwing out three-quarters of patches ($\alpha=0.25$).

Some patches that are irrelative to the object in an image are unpredictable, such as the background.
A large percentage of patches are trivial, and we could fast pre-train our model by simply removing them.

\subsection{Transferability}
To study the transferability of intra-object semantics, we use COCO with high-quality segmentation for pre-training. Then, we evaluate the pre-trained model on three downstream tasks: fine-tuning on ImageNet-100, linear probing on ImageNet-100, fine-tuning on CIFAR100.

% \usepackage{booktabs}
% \usepackage{booktabs}
\begin{table*}[]
\centering
\begin{tabular}{lll|ll|ll|l}
\toprule
       &          &     & \multicolumn{2}{l|}{Fine-tune}           & \multicolumn{2}{l|}{Linear probe} & CIFAR100 \\
Model  & $\alpha$ & Speed& Top-1                    & Top-5       & Top-1         & Top-5        & Top-1    \\ \midrule
ViT-S  & 1.00     & 1.0x   &{73.20}                  &  {89.72}    &-              & -           &  60.02          \\
MAE    & 0.50     & 1.9x     & 89.52                    & 98.16       & 33.30         & 60.74        & 82.90         \\
ObjMAE & 0.34     & 2.1x     & 87.52                    & 97.48       & 29.84         & 57.16        &   81.80    \\
\bottomrule
\end{tabular}
\caption{Performance on ImageNet100 of models pre-trained on COCO. Note that, ViT-S is not a pre-trained model.}\label{tab:perf_coco}
\end{table*}

% \begin{table}[]
% \centering
% \begin{tabular}{@{}llrlr@{}}
% \toprule
%       & \multicolumn{2}{c}{finetune}                          & \multicolumn{2}{c}{linprobe}                     \\ 
% Model  & Top-1                     & \multicolumn{1}{l}{Top-5} & Top-1                & \multicolumn{1}{l}{Top-5} \\ \midrule
% MAE    &                           &                           & \multicolumn{1}{r}{} &                           \\
% ObjMAE & \multicolumn{1}{r}{89.02} & 97.98                     & \multicolumn{1}{r}{} &                           \\ \bottomrule
% \end{tabular}
% \caption{Performance on ImageNet100 pre-trained on ImageNet100.}\label{tab:perf_imagenet100}
% \end{table}
We compare the performances of MAE, ViT, and ObjMAE in Table~\ref{tab:perf_coco}.
The pre-trained models outperform the ones without pre-training for all benchmarks.
ObjMAE remains 90\%-98\% performances with removing 83\% visual patches.

\begin{figure}
    \centering
    \includegraphics[width=.9\linewidth]{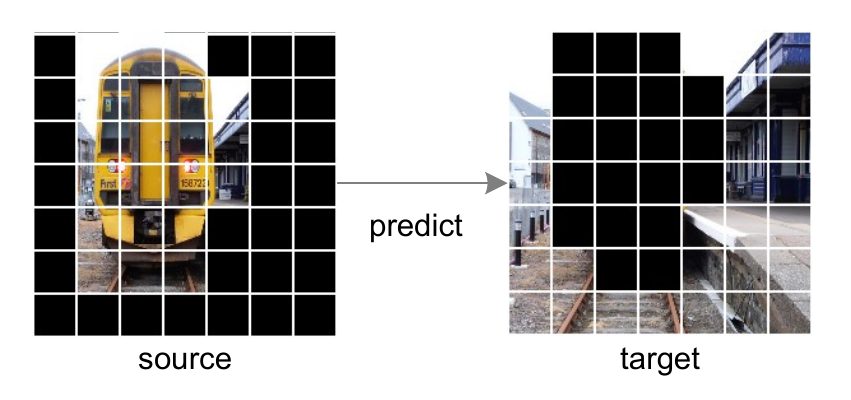}
    \vspace{-0.3cm}
    \caption{Inter-object prediction task. We randomly select two different object and choose one for the source and the other for target. Sometimes, there is only one object in the image, so we regard the background as one object. Usually, the background is not an object.} \label{fig:inter-object_task}
\end{figure}

\begin{figure}
    \centering
    \includegraphics[width=\linewidth]{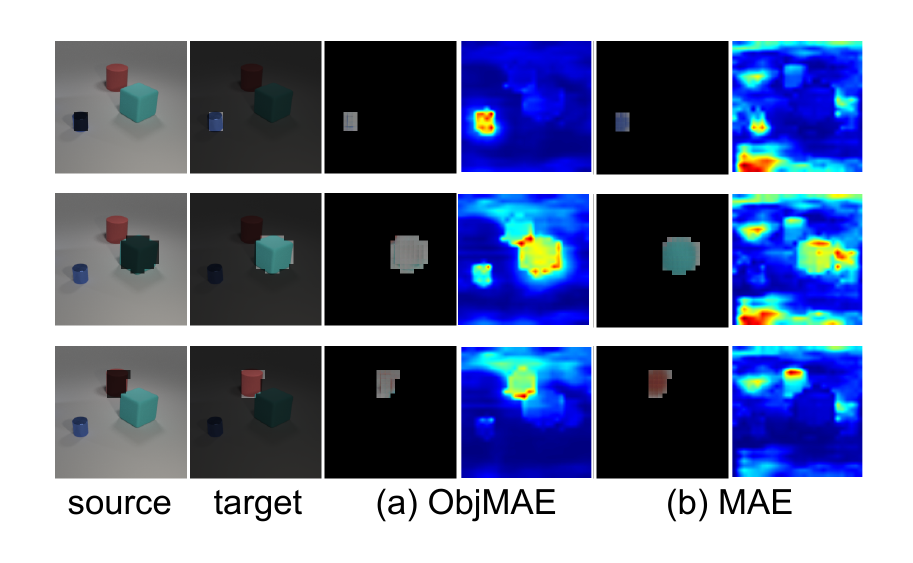}
    \vspace{-0.9cm}
    \caption{Attention rollout~\citep{rollout} and reconstruction for ObjMAE and MAE. A region belonging to an object is chosen as the target region and the rest belongs to the source. The unselected region is darker than the chosen one. (a) ObjMAE fails to predict the unseen object, and the object region merely responses to the other regions. (b) MAE can predict the unseen object, and the object region responses to the other regions.}
    \label{fig:representation_comparision}
\end{figure}

\subsection{Inter-object and Intra-object Semantics}
% \usepackage{booktabs}
\begin{table}[]
\centering
\begin{tabular}{@{}lll@{}}
\toprule
Selective Region & Top-1         & Top-5          \\ \midrule
Intra-object     & \textbf{88.4} & \textbf{97.64} \\
Inter-object     & 86.34         & 97.06          \\
\bottomrule
\end{tabular}
\caption{The performance on ImageNet100 of models pre-trained on COCO  for selective regions.}\label{tab:inter-intra}
\end{table}
There are two kinds of semantics on an image: inter-object and intra-object semantics.
ObjMAE only conducts the predication task among one object, learning the intra-object semantics.
MAE randomly picks patches from the whole image including both inter-object and intra-object semantics.
To further study the role of intra-object semantics on pre-training models, we introduce a new selection strategy to select the source region from one object and select the target region from another object as shown in Figure~\ref{fig:inter-object_task}.

Table~\ref{tab:inter-intra} shows the results for the strategies about the selective regions.
Experimental results show that the model learning intra-object semantics outperforms the one learning inter-object semantics.
It suggests that intra-object semantics is more important than inter-object semantics.

\subsection{Patch size}
% no constructive conclusion here, so it is considerable to remove.
% \usepackage{booktabs}
\begin{table}[]
\centering
\begin{tabular}{@{}lll|ll@{}}
\toprule
Model & $\alpha$ &Patch Size & Top-1         & Top-5          \\ \midrule
MAE   & 1    &   16      & 89.52	        & 98.16 \\
MAE   & 1    &   8       & 82.28	        & 95.68         \\
ObjMAE & 0.34&   8       & 82.14	        & 95.58         \\
\bottomrule
\end{tabular}
\caption{The performance on ImageNet100 of models pre-trained on COCO  for patch size. }\label{tab:patch_size}
\end{table}
Could the performance of the pre-training model be further improved by reducing the size of the patch?
ObjMAE makes it possible to train a pre-training model with a small patch size by a small object ratio.
Unfortunately, the results in Table~\ref{tab:patch_size} demonstrate a significant decay in performance. 

\subsection{Representation Visualization}

\begin{figure}[!htb]
    \centering
    \includegraphics[width=.9\linewidth]{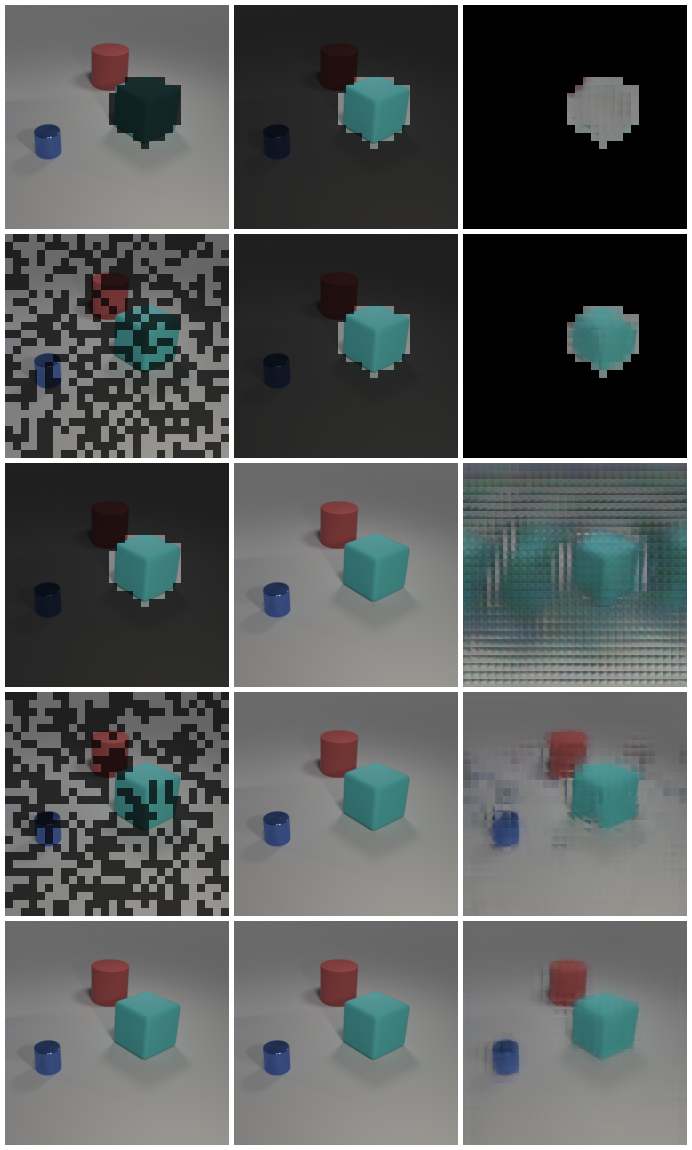}
    \caption{Target reconstruction for the source patches. From left to the right, the first column is the source patches, the second column is the target patches, and the third column is the reconstruction patches for the target patches. The masked patches are darker.}
    \label{fig:target_reconstruction}
\end{figure}

We train ObjMAE ($\alpha=0.05$) and standard MAE on CLEVR-M for 300 epochs.
In Figure~\ref{fig:representation_comparision}, we visualize the attention rollout~\citep{rollout} and reconstruction for predicting one object through other regions on the training set. 
ObjMAE cannot reconstruct the object without seeing any patch of it, while MAE rebuilds it successfully.
It is a marvel that MAE can rebuild the unseen object, indicating that MAE memorizes the whole scene and representation. 
In contrast, ObjMAE can only infer the inter-object information, which learns an object-wise representation.
The attention rollout shows that ObjMAE focuses on an object and learns intra-object semantics.
In contrast, MAE captures a global representation and learns both inter-object semantics and inter-object semantics.

We demonstrate more results of ObjMAE for the choices of the source patches and the target patches in Figure~\ref{fig:target_reconstruction}.
Experimental results show that ObjMAE can correctly predict an object (the cube) only if it has seen the object.
Though it only learns the intra-object semantics, ObjMAE still has the ability to imagine the surrounding objects, see in the third row.
Such an imagination does not violate ObjMAE to reconstruct the real objects (row 4 \& 5).

\section{Ablation Studies}
We conduct comprehensive and ablative studies to understand the difference between the padding modes---zero padding and replicate padding.
We also experiment with a modified MAE that randomly throws out patches to study the effect of object ratio.

\subsection{Padding Mode}
In this part, we compare padding modes of ObjMAE. 
We first train models with different padding modes for 300 epochs on COCO to obtain pre-training models and then fine-tune them for 100 epochs on ImageNet-100 to evaluate their performances.
The results are summarized in Table~\ref{tab:padding_mode}.
Random padding could learn the best model, but the improvement is insignificant (less than 2\%).

% \usepackage{booktabs}
\begin{table}[!htb]
\centering
\begin{tabular}{@{}lll@{}}
\toprule
Padding Mode & Top-1         & Top-5          \\ \midrule
Zero         & 83.38	     & 94.16 \\
Replicate    & 82.58	     & 94.04          \\
Random       & 83.48	     & 93.78      \\
\bottomrule
\end{tabular}
\caption{Ablation study for padding mode. The results are pre-trained on COCO and fine-tuned on ImageNet-100. The maximum error is 2\% or less.}\label{tab:padding_mode}
\end{table}

\begin{table}[!htb]
\centering
% \begin{tabular}{l|lll|lrr}
% \toprule
%       & \multicolumn{3}{c|}{COCO} & \multicolumn{3}{c}{ImageNet-100}                     \\
% Model & $\alpha$ & Top-1 & Top-5 & $\alpha$ & \multicolumn{1}{l}{Top-1} & \multicolumn{1}{l}{Top-5} \\ \midrule
% ObjMAE & 0.25    & 87.52  & 97.48  & 0.25  & 88.69                 & 98.06                 \\
% ObjMAE & 0.17   & 88.40   & 97.64  & 0.21  & 89.02                 & 97.98                 \\
% ObjMAE & 0.05    & 86.6   & 97.22  & 0.05  & \multicolumn{1}{l}{-} & \multicolumn{1}{l}{-} \\ \midrule
% \midrule
% MAE    & 0.25    & 87.52  & 97.48  & 0.25  & 89.32                 & 98.26                 \\
% MAE    & 0.17   & 88.38   & 97.64  & 0.21 & 89.58                 & 98.08                 \\
% % MAE    & 0.1    & 86.6   & 97.22  & 0.1  & \multicolumn{1}{l}{-} &\multicolumn{1}{l}{-} \\
% \bottomrule
% \end{tabular}

\begin{tabular}{ll|ll}
\toprule
    %   & &\multicolumn{2}{c}{COCO} \\
Model  & $\alpha$  & Top-1 & Top-5 \\ \midrule
ObjMAE & 0.25      & 87.52 & 97.48 \\
ObjMAE & 0.17      & \textbf{88.40} & 97.64 \\
ObjMAE & 0.05      & 86.6  & 97.22 \\ \midrule
MAE    & 0.50      & 89.52 & 98.16 \\
MAE    & 0.25      & 88.86 & 97.72 \\
MAE    & 0.17      & 88.38 & 97.62 \\ 
\bottomrule
\end{tabular}
\caption{Ablation study for $\alpha$. The results are pre-trained on COCO and fine-tuned on ImageNet-100. ObjMAE beats MAE when $\alpha=0.17$. }\label{tab:obj_ratio}
\end{table}

\subsection{Object Ratio}
The model can be faster pre-trained by reducing $\alpha$.
However, a small value of $\alpha$ may diminish the performance for throwing important information.
To study the effect of $\alpha$, we also set $\alpha$ for MAE to control the length of the source patches and the target patches.
MAE ($\alpha=0.5$) is the standard MAE with a masking ratio of 0.5, which takes half of the patches as the input.

The experimental results in Table~\ref{tab:obj_ratio} demonstrate that The performance of MAE has a downtrend as reducing $\alpha$, and ObjMAE gets the best result when the number of utilized patches both source and target meets the average object size (0.17 is half of the average object ratio of COCO).
Object-wise selection becomes vital to performance when the visible patches are limited, and the intra-object patches play a more critical role in pre-training than the inter-object patches.

\section{Conclusion}

In this work, we present ObjMAE with a light self-supervised prediction task for only the object region, which ignores the inter-object semantics.
Specifically, we introduce the object selection and division strategy to drop non-object patches for learning object-wise representations by selective reconstruction with interested region masks.
Extensive experiments on three commonly-used datasets demonstrate the efficiency of learning the intra-object semantics compared with learning from the whole image.
Furthermore, the intra-object prediction is more effective than the inter-object prediction for self-supervised pre-training.

\section{Discussions}

In this work, we verify the importance of intra-object semantics in self-supervised training and show that the effect of inter-object semantics is much more complex.
On the one side, it may interfere with the model to make wrong decisions by learning biases on the data. For example, fish and water always appear in couples, which misleads the model to increase the confidence of categorizing an object to fish when seeing water.
On the other hand, such inter-object semantics provides clues to infer the object when it is hard to distinguish.
For future work, we plan to enrich this study on the following aspects:

\begin{itemize}
    \item[] 1) We only use a part of COCO, the training set of COCO2014. Actually, there are over 330K images available, which makes it possible to build a large-scale dataset for object-wise pre-training.
    \item[] 2) It makes sense to propose an end-to-end object-wise representation method by combining an unsupervised segmentation method.
    \item[] 3) More downstream tasks are considered to evaluate the performance of object-wise representation, such as segmentation and visual reasoning. In particular, visual reasoning inferring the relationships of objects may benefit from removing the biases learning from inter-object semantics on the training data.
    \item[] 4) Most objects on COCO are too small to be separated into enough patches for the object-wise prediction tasks. For now, we simply discard these objects which contain more subtle information about the image. Hopefully, the performance of ObjMAE will be improved by utilizing these small objects.
\end{itemize}


\clearpage

\bibliography{references}
\bibliographystyle{icml2022}

\clearpage

%%%%%%%%%%%%%%%%%%%%%%%%%%%%%%%%%%%%%%%%%%%%%%%%%%%%%%%%%%%%%%%%%%%%%%%%%%%%%%%
%%%%%%%%%%%%%%%%%%%%%%%%%%%%%%%%%%%%%%%%%%%%%%%%%%%%%%%%%%%%%%%%%%%%%%%%%%%%%%%

\appendix

\onecolumn

\section*{Appendix}

\section{Implementations Details about Selection with CAM}
The class activation map $\mathcal{M}_c$ for class $c$ can be calculated by
\begin{equation}
    \mathcal{M}_c(x,y) = \sum_k w_k^c f_k(x,y),
\end{equation}
where $f_k(x,y)$ represents the activation of unit $k$ in the last convolutional layer of the pre-trained ResNet-50~\citep{resnet} at spatial location $(x, y)$.
Then, $\mathcal{M}_c(x,y)$ is normalized by dividing the maximal value on $\mathcal{M}_c(x,y)$ into a range of 0-1.
The pixels with $\mathcal{M}_c(x,y)$ larger than 0.3 are regarded as a part of object region.

\begin{figure}[!htb]
    \centering
    \includegraphics[width=.7\linewidth]{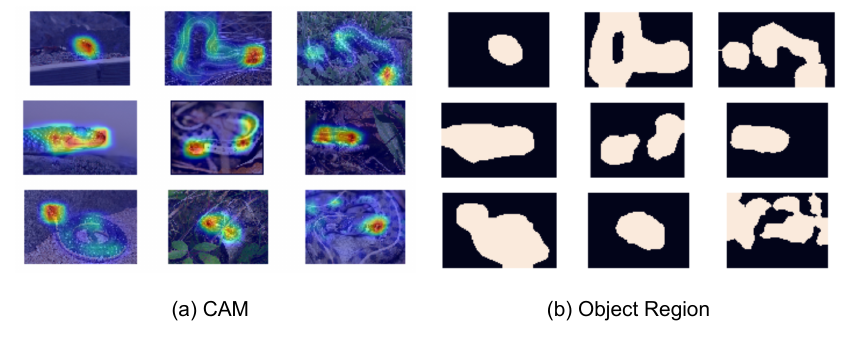}
    \vspace{-0.3cm}
    \caption{Segmentation by CAM. Threshold for selecting the object region is 0.3.}
    \label{fig:seg_cam}
\end{figure}

We show the samples on ImageNet-100 and their segmentation by CAM in Figure~\ref{fig:seg_cam}.

\section{Visualizations of Object-based Attention}

\begin{figure}[!htb]
    \centering
    \includegraphics{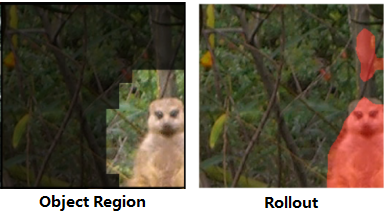}
    \caption{Object-based attention. The left light area is the object region. The right image shows the segmentation of the object region by rollout. Our model refines the object region.}
    \label{fig:object-based_attention}
\end{figure}

We use rollout~\citep{rollout} to visualize the attention heatmap of ObjMAE ($\alpha=0.17$) for the object region in Figure~\ref{fig:object-based_attention}.
Though the selective region of an animal by CAM is wider than the actual region, ObjMAE captures a more precious region for the animal.
Our model regards the tree trunk on the background as a part of the animal, because they have a similar texture.

\end{document}